\documentclass[a4paper, 10pt, conference]{ieeeconf}

\IEEEoverridecommandlockouts %

\usepackage{graphicx}
\usepackage{booktabs}
\usepackage{amsmath,amssymb}
\usepackage[hidelinks]{hyperref}

\newcommand{\pizerofive}{$\pi_{0.5}$}

\title{\LARGE \bf
MINERVA: How Small Can a Manipulation Policy Be and Still Solve LIBERO?
}

\author{Kohei Sendai$^{1}$, Tatsuya Matsushima$^{2}$, Yusuke Iwasawa$^{3}$%
\thanks{$^{1}$Kohei Sendai is with Matsuo-Iwasawa Lab, Graduate School of Engineering,
        The University of Tokyo, Japan
        {\tt\small kohei.sendai@weblab.t.u-tokyo.ac.jp}}%
\thanks{$^{2}$Tatsuya Matsushima is with Matsuo-Iwasawa Lab, The University of Tokyo,
        {\tt\small matsushima@weblab.t.u-tokyo.ac.jp}}%
\thanks{$^{3}$Yusuke Iwasawa is with Matsuo-Iwasawa Lab, The University of Tokyo,  {\tt\small iwasawa@weblab.t.u-tokyo.ac.jp}}%
}

\IEEEaftertitletext{%
\vspace{0.2\baselineskip}
\begin{center}
\includegraphics[width=\textwidth]{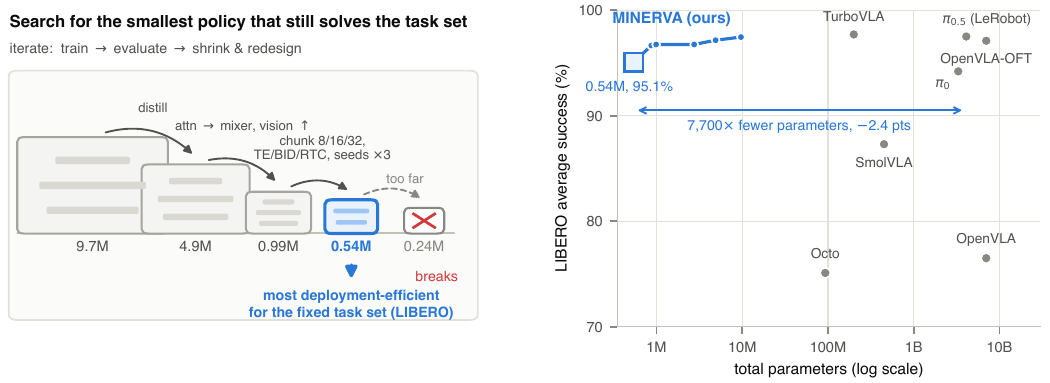}\\[3pt]
\parbox{0.97\textwidth}{\refstepcounter{figure}\label{fig:teaser}%
\footnotesize Fig.~\thefigure.\quad \textbf{Left:} we search for the
smallest policy that still solves the fixed LIBERO task set, iterating over
model scale, capacity allocation, objectives and inference strategies; the
search stops at a 0.54M-parameter empirical capacity floor --- one step
smaller and the benchmark breaks.
\textbf{Right:} that floor (blue) lands in the same roughly 95--98\% band as
fine-tuned generalist VLAs two to four orders of magnitude larger (gray,
paper-reported: Octo~\cite{octo}, OpenVLA~\cite{openvla},
SmolVLA~\cite{smolvla}, $\pi_0$~\cite{pi0}, TurboVLA~\cite{turbovla},
OpenVLA-OFT~\cite{oft}, as compiled in~\cite{smolvla,turbovla,oft});
the \pizerofive{} point is the LeRobot implementation's reported
result~\cite{pi05,lerobot}.}
\end{center}
\vspace{0.6\baselineskip}}

\begin{document}

\maketitle
\thispagestyle{empty}
\pagestyle{empty}

\begin{abstract}
Vision-language-action (VLA) models with billions of parameters now dominate the LIBERO manipulation benchmark, yet how much capacity the benchmark actually demands has not been measured. From a deployment standpoint this is the quantity that matters: once a robot's task set is fixed, the smallest policy that still solves it governs deployment cost --- for example, whether the policy fits on an edge device. We present MINERVA (MINimal Efficient Robotic Vision-Action policy), a family of deliberately minimal visuomotor policies and scale it down until it breaks. A 0.54M-parameter policy reaches 95.1\% average success over 2,000 rollouts on the four standard LIBERO suites, 2.4 points below the reported LeRobot $\pi_{0.5}$ result from a model 7,700$\times$ larger; performance saturates near one million parameters and collapses below a quarter million. Across a broad sweep of architectural, training, and inference choices, only action-chunk length and vision allocation consistently exceed a $\pm$1-point training-seed band. Flow matching shows no detectable advantage over direct L1 regression across three seeds, while regression is up to 3.8$\times$ faster on GPU. A permutation probe makes the role of task conditioning causal: rewriting the task-ID mapping alone collapses success to near chance, showing that instruction conditioning on standard LIBERO primarily selects among memorized tasks. The same recipe reaches 94.6\% over 89 LIBERO-90 tasks in a single-seed extension, while LIBERO-Plus evaluation drops success to 46--56\% under perturbations, with near-zero photometric robustness across all tested scales. The 0.54M policy replans at every control step in 5--9 ms per chunk on a laptop CPU --- 113$\times$ faster than SmolVLA and 1,400$\times$ faster than \pizerofive{} --- with no GPU. We position this study as a first step toward measuring task-specific capacity floors, which can guide the construction or distillation of deployment-efficient policies.
Code, recipes and all checkpoints are open sourced here:
\url{https://github.com/k1000dai/MINERVA}.
\end{abstract}

\section{INTRODUCTION}

Imitation-learned manipulation policies are growing rapidly:
vision-language-action (VLA) models such as OpenVLA (7B)~\cite{openvla},
$\pi_0$~\cite{pi0} and \pizerofive{}~\cite{pi05} (3--4B) attach action heads
to pretrained vision-language backbones and report near-saturated success on
the LIBERO benchmark~\cite{libero}. The costs follow the size: multi-GPU
training, gigabytes of weights, large VRAM for inference. For system integrators who need a manipulation module on a CPU, beside a planner, under a power budget, the practical question is not
whether a larger model helps, but how much capacity the task demands.

This is a deployment-efficiency question in the sense of~\cite{bremen}:
optimize for the constraints that bind at deployment rather than during
development. Generalization is what a large VLA buys, and it matters --- but
it is not required everywhere. Once a robot is deployed its task set is
typically fixed, and what then governs cost is the task set's \emph{minimal
sufficient capacity}: the smallest parameter count at which it is still
solved. If that capacity floor is known, a policy can be built at it ---
directly, or by distilling a generalist~\cite{hinton} --- and run on the
edge. Conceptually, such a floor is a property of the task setting, in the
spirit of
information-theoretic task-complexity measures in RL~\cite{pic}, and it
will differ across environments, sensing noise and embodiments; but it has
to be measured somewhere first. In this work, we use \emph{empirical
capacity floor} to denote the smallest tested model in our policy family
that maintains the target success level under the fixed training and
evaluation protocol.

This paper measures that demand for the most widely used benchmark in the
field. Recent analyses~\cite{liberopro} show that standard LIBERO evaluation
is largely a memorization test: policies are trained and evaluated on the same
40 tasks in the same scenes, and the language instruction mostly serves to
disambiguate which of the memorized tasks to execute. We take this
observation seriously and build MINERVA (MINimal Efficient Robotic
Vision-Action policy), a policy family designed to contain
nothing beyond what that reading of the benchmark requires: a from-scratch CNN
encoder (no pretrained backbone), a 40-entry learned task-ID embedding (no
language encoder), and a flow-matching action-chunk head whose token mixer is
an MLP rather than self-attention. We then scale the total budget from 9.7M
parameters down to 0.09M and observe where performance breaks.

The headline result is that \textbf{0.54M parameters reach 95.1\% average
success} over the four standard suites, 2.4 points below the reported
LeRobot \pizerofive{} result from a model 7{,}700$\times$ larger; 0.99M
parameters
land within 0.75 points (Table~\ref{tab:headline}, Fig.~\ref{fig:teaser}).
Success saturates near 1M
parameters and collapses below 0.25M --- and the collapse consistently arrives
through the long-horizon suite first, while short-horizon tasks survive in
models far too small to chain subgoals (Fig.~\ref{fig:scaling}).

The small parameter count makes broad design-space exploration cheap enough to be practical. We sweep architectural and training choices including action-chunk length, vision--action capacity allocation, generative versus regression objectives, and action-token mixing, and re-train the key ablation configurations under three seeds. This reveals a $\pm$1-point seed band on the 4-suite average --- comparable to many single-run ablation deltas on this benchmark --- and leaves only two robust effects: action-chunk length ($-3.2$ points at chunk 8, $-2.2$ at chunk 32), with short chunks breaking the goal suite and long chunks the long-horizon suite, and vision allocation, where starving the encoder costs a seed-replicated $-1.41$ points. By contrast, flow matching shows no detectable advantage over direct L1 regression across three seeds ($+0.34$, inside the band), while regression requires one forward pass instead of ten Euler steps. Self-attention over the 16 action tokens is likewise unnecessary: a token-mixing MLP matches its score with 26\% fewer parameters, and reallocating that capacity to vision is what enables the sub-1M models.

Three further studies delimit what this level of performance is and is not:
a \emph{permutation probe} isolates the causal role of task conditioning (rewriting
the task-ID mapping alone collapses success from 96.75\% to chance level); the
recipe survives $2.25\times$ the task count (94.6\% on LIBERO-90 at 0.995M);
and a \emph{LIBERO-Plus} audit prices what the headline omits (46--56\%
under perturbation; photometric robustness remains near zero at every
tested scale).

Inference-time strategy matters as much as architecture. Replanning every step
and averaging overlapping chunk predictions (ACT-style temporal
ensembling~\cite{act}) outperforms both BID-style candidate selection~\cite{bid} and soft-inpainting (RTC~\cite{rtc}) approaches, and
mode-seeking sampling (initial-noise temperature 0.85) adds up to 2.3 points
--- but only on sub-1M models, where the learned velocity field is noisiest.
The resulting policies are lightweight enough for CPU deployment: 8.9\,ms per action chunk on
eight laptop CPU threads for the 0.54M model, against 1.0\,s for
SmolVLA~\cite{smolvla} and 12.8\,s for \pizerofive{}
(Table~\ref{tab:speed}).

Contributions:
(i) an empirical capacity floor for standard LIBERO: 0.5M parameters, no
language encoder and no pretrained perception suffice for 95\%, with a
parameter-scaling curve localizing where capacity stops mattering;
(ii) a seed-replicated ablation exposing a $\pm$1-point training-seed band on
this benchmark and showing that only chunk length and vision allocation
survive it, while flow matching shows no detectable advantage over one-pass regression;
(iii) a budget-allocation study showing vision, not the action head, is where
parameters pay;
(iv) a comparison of inference-time chunking strategies under one
protocol; and
(v) a permutation probe of task-ID conditioning, a LIBERO-90 extension, and a
LIBERO-Plus robustness audit that together delimit what a
memorization-level policy does and does not achieve.

\section{RELATED WORK}

\textbf{VLA policies and LIBERO.}
LIBERO~\cite{libero} provides four ten-task suites (Spatial, Object, Goal,
Long/``10'') and paired demonstrations, and has become the default benchmark
for VLA evaluation. Current leaders are large: OpenVLA (7B)~\cite{openvla}
fine-tunes a Prismatic VLM, and OpenVLA-OFT~\cite{oft} lifts it to 97.1\%
with an optimized fine-tuning recipe; $\pi_0$~\cite{pi0} and
\pizerofive{}~\cite{pi05} attach flow-matching action experts to a PaliGemma
backbone; Octo (93M)~\cite{octo} and SmolVLA (450M)~\cite{smolvla} are the
efficiency-oriented entries, still two to three orders of magnitude larger
than our 0.54M model. LIBERO-PRO~\cite{liberopro} shows
these scores largely measure memorization --- perturbing objects, layouts or
instructions collapses standard models to near 0\% --- which motivates our
design: if standard LIBERO is a memorization benchmark, its capacity floor is
an empirical question, and language should be replaceable by a task index.

\textbf{Efficient VLAs.}
A growing body of work attacks VLA inference cost while keeping the
vision-language backbone: TinyVLA~\cite{tinyvla} pairs a sub-1B VLM with a
diffusion head and skips generalist pretraining; MiniVLA~\cite{minivla} swaps
OpenVLA's 7B backbone for a 1B VLM; DeeR-VLA~\cite{deer} adds dynamic
early-exit inference; FAST~\cite{fast} compresses the action token stream;
SmolVLA~\cite{smolvla} trims the recipe to 450M. Closest in spirit to us,
TurboVLA~\cite{turbovla} removes the large language model from the control
pathway entirely and reaches 97.7\% on LIBERO with 0.2B parameters at
32\,Hz on an RTX~4090. All of these retain a language pathway and pretrained
components, and none goes below 0.2B. MINERVA asks the complementary
question --- not how to make a VLA cheaper, but how much capacity the
benchmark itself demands --- and its headline model lands more than two
orders of magnitude below the smallest of them, at the declared cost of any
open-vocabulary capability. The framing follows deployment-efficient
RL~\cite{bremen}, which optimizes for the constraints that bind at
deployment, and task-complexity measurement~\cite{pic}, which quantifies
what a task demands independently of any particular algorithm.

\textbf{Small visuomotor policies.}
Sub-100M imitation policies predate the VLA wave: ACT~\cite{act} (an 80M
transformer with action chunking and temporal ensembling) and Diffusion
Policy~\cite{diffusionpolicy} remain strong baselines, and end-to-end
CNN policies with spatial-softmax keypoints~\cite{levine} date to 2016. MINERVA
inherits from all three --- chunking and ensembling from ACT, a generative
action head from Diffusion Policy (flow matching~\cite{flowmatching} instead
of DDPM), keypoint extraction from~\cite{levine} --- but pushes total capacity
one to two orders of magnitude lower than prior work while remaining
multi-task across 40 tasks.

\textbf{Efficient generative heads.}
Our head follows DiT \cite{dit} with AdaLN-Zero conditioning and shares the AdaLN projection across blocks. Replacing token
self-attention with an MLP mixer follows~\cite{mlpmixer}; depthwise-separable
convolutions follow MobileNets~\cite{mobilenet}; FiLM
conditioning~\cite{film} injects the task embedding into the encoder.
Distillation~\cite{hinton} transfers velocity targets from a larger teacher.

\textbf{Inference-time chunking strategies.}
How a chunked policy is \emph{executed} is an active design axis: temporal
ensembling averages overlapping predictions~\cite{act}; Bidirectional Decoding
(BID)~\cite{bid} re-samples candidate chunks and selects for coherence with
the committed plan; Real-Time Chunking (RTC)~\cite{rtc} softly inpaints the
new chunk toward the old plan during flow integration. We evaluate all three,
plus initial-noise temperature control, on one model under one protocol.

\section{METHOD}

\subsection{Design principle: nothing the benchmark does not demand}

Standard LIBERO evaluates on the training scenes, tasks and instructions. If the instruction is used only to identify one of the 40 known tasks, its functional role can be represented by a 40-way task ID. MINERVA replaces the language pathway with a
learned embedding table of 40 entries, and spends the entire parameter budget
on the two things the benchmark does exercise: visual state estimation and
closed-loop action generation. This is explicitly a study of standard LIBERO,
not a general-purpose VLA; Section~\ref{sec:limitations} returns to what is
given up.

\subsection{Architecture}

\textbf{Perception.}
Both camera views (agent-view and wrist, $256^2$, resized to $144^2$ and
random-cropped to $128^2$ at training, center-cropped at evaluation) pass
through one shared from-scratch CNN with a learned view embedding:
four stride-2 stages of depthwise-separable residual blocks~\cite{mobilenet},
FiLM-modulated~\cite{film} by the task embedding, followed by spatial-softmax
keypoint extraction~\cite{levine} and a linear projection to a per-view
feature. A two-layer MLP encodes the 8-D proprioceptive state. The
concatenation of view features, state features and the task embedding,
over $n_{\text{obs}}{=}2$ observation steps, forms the conditioning vector
$c$. Observation history matters: at 7M parameters, moving from one to two
observation steps gains $+6$ points on Goal and $+4$ on Long.

\textbf{Action head.}
Actions are generated as chunks $a = a_{1:H}$ with $H{=}16$ by conditional
flow matching~\cite{flowmatching}. Training draws $\varepsilon \sim
\mathcal{N}(0, I)$ and a noise level $\tau$, forms the interpolant $a^\tau =
\tau a + (1-\tau)\varepsilon$, and regresses the velocity field toward the
constant target:
\begin{equation}
\mathcal{L}_{\mathrm{fm}} =
\mathbb{E}_{a,\varepsilon,\tau}
\left\| v_\theta(a^\tau, \tau, c) - (a - \varepsilon) \right\|_2^2 ,
\end{equation}
with $\tau$ sampled from a Beta distribution tilted toward high noise. At
inference, a chunk is produced by integrating $v_\theta$ with 10 Euler steps
from $\varepsilon \sim \mathcal{N}(0, \sigma^2 I)$, where $\sigma$ is the
noise temperature of Section~\ref{sec:method-inference}. The head is a
4-layer DiT~\cite{dit} in which self-attention over the 16 action tokens is
replaced by a token-mixing MLP~\cite{mlpmixer}, and the AdaLN-Zero modulation
MLP is shared across blocks~\cite{ditair}. The same head can instead be
trained as a direct L1 regression (learned queries in, chunk out, one forward
pass at inference); Section~\ref{sec:seeds} shows flow matching has no detectable advantage over direct L1 regression across three seeds on this benchmark. A training-only auxiliary head predicts episode progress ($<$9k parameters, dropped at inference).

\textbf{Model family.}
Only the two sub-1M models differ purely in width: MINERVA-0.5M (0.54M:
CNN widths 32/64/112/160, 48 keypoints, head width 96) and MINERVA-1M
(0.99M: 40/80/160/224, 64 keypoints, head width 128); MINERVA-5M (4.89M)
restores the full-width CNN and head self-attention, and MINERVA-10M
(9.66M) also unshares AdaLN. Each smaller model is trained with velocity
distillation~\cite{hinton} from a larger teacher: a frozen teacher
$v_\phi$ is evaluated at the same $(a^\tau, \tau, c)$ and its prediction
added as a second regression target,
$\mathcal{L} = \mathcal{L}_{\mathrm{fm}} +
\lambda \| v_\theta - v_\phi \|_2^2$ with $\lambda{=}1$.
The teacher is excluded from parameter counts and absent at inference. A
single-run comparison at 5M favored distillation strongly ($+1.85$ average
over the same architecture trained alone; the distilled 4.89M model even
beats its own 7.12M teacher), while at 1M it sits inside the seed band of
Section~\ref{sec:seeds} --- we keep it for the mid-scale models.

\subsection{Inference-time strategy}
\label{sec:method-inference}

At deployment the policy replans every control step and applies ACT-style
temporal ensembling~\cite{act}: each executed action is the exponentially
weighted average (decay $0.01$ per step of age) of the predictions from all
chunks covering that timestep. For sub-1M models the initial noise of the
flow integration is scaled by $\sigma = 0.85 < 1$, making sampling
mode-seeking. Both choices are validated in Section~\ref{sec:inference}.

\section{EXPERIMENTAL SETUP}
\label{sec:setup}

\begin{figure}[tb]
\centering
\includegraphics[width=\linewidth]{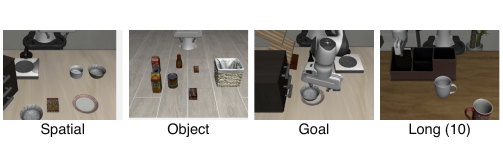}
\caption{The four standard LIBERO task suites: Spatial, Object, Goal  and Long/``10''. Scene renderings from the LIBERO benchmark~\cite{libero}.}
\label{fig:tasks}
\end{figure}

\textbf{Data and training.}
All models train on the \texttt{lerobot/libero} dataset (1{,}693
demonstrations, 273k frames, all 40 tasks pooled across the four
suites; Fig.~\ref{fig:tasks}),
implemented in the LeRobot framework~\cite{lerobot}. Training runs 90--120k
steps at batch 128 with AdamW (lr $10^{-4}$, cosine decay); one model trains
in $\approx$2\,h on a single GH200 or consumer RTX~5080 GPU. Long-suite
episodes are loss-upweighted $3\times$ for the sub-1M models.

\textbf{Evaluation protocol.}
Unless otherwise noted, standard-LIBERO results use 50 episodes per task
$\times$ 10 tasks $\times$ 4 suites $=$ 2{,}000 rollouts under fixed
seed, with hard environment resets between episodes. 
Ablation screens use 25 episodes per task where noted; all
conclusions were re-confirmed at 50.

\section{RESULTS}

\subsection{Headline: 95\% at half a million parameters}

\begin{table}[tb]
\caption{Success rate (\%) on the four LIBERO suites. MINERVA rows use
2{,}000 rollouts (50 episodes/task), fixed seed.
The \pizerofive{} row is the LeRobot implementation's reported
result~\cite{lerobot} (10 episodes/task, 400 rollouts).}
\label{tab:headline}
\centering
\small
\setlength{\tabcolsep}{4.5pt}
\begin{tabular}{@{}lrrrrrr@{}}
\toprule
policy & params & Spat. & Obj. & Goal & Long & \textbf{avg} \\
\midrule
MINERVA-0.5M & 0.54M & 94.4 & 99.6 & 96.4 & 89.8 & \textbf{95.05} \\
\quad + L1 head & 0.54M & 96.8 & 99.6 & 97.4 & 89.2 & 95.75 \\
MINERVA-1M   & 0.99M & 97.0 & 99.8 & 97.4 & 92.8 & 96.75 \\
MINERVA-5M   & 4.89M & 99.0 & 99.2 & 98.0 & 92.4 & 97.15 \\
MINERVA-10M  & 9.66M & 98.4 & 99.0 & 98.4 & 94.0 & 97.45 \\
\midrule
\pizerofive{} (LeRobot) & 4.1B & 97.0 & 99.0 & 98.0 & 96.0 & 97.50 \\
\bottomrule
\end{tabular}
\end{table}

Table~\ref{tab:headline} shows the released family against the LeRobot
implementation of \pizerofive{}, whose reported result~\cite{lerobot} was
measured at 10 episodes/task --- a reference point rather than a
protocol-matched comparison. The 0.54M model reaches
95.05\% --- 2.4 points behind a model 7{,}700$\times$ its size --- and the
0.99M model closes to 0.75 points. Retrained with the one-pass regression
head of Section~\ref{sec:seeds}, the same 0.54M architecture scores 95.75\%;
this is the configuration we ship. The remaining gap is almost entirely the
long-horizon suite (92.8 vs.\ 96.0 at 1M); the three short-horizon suites are
matched or exceeded from 1M upward. Rows are single runs (seed 1000);
Section~\ref{sec:seeds} re-trains the 1M model under three seeds
(94.60--96.75) and reads every comparison against that band.

\subsection{Scaling: saturation at 1M, collapse below 0.25M}

\begin{figure}[tb]
\centering
\includegraphics[width=\linewidth]{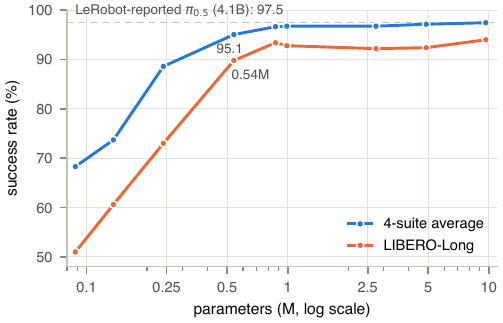}
\caption{Success versus total parameters (log scale), 2{,}000 rollouts per
point. Performance saturates near 1M parameters; below 0.25M it collapses,
and the long-horizon suite (orange) falls first.}
\label{fig:scaling}
\end{figure}

Fig.~\ref{fig:scaling} extends the family across two orders of magnitude.
From 1M to 10M parameters, average success moves only 0.7 points
(96.75$\to$97.45): capacity stops being the binding constraint at about one
million parameters. Below 0.54M the curve breaks sharply --- 88.6\% at 0.24M,
68.3\% at 0.09M --- and the failure is structured: at 0.24M the model still
solves 95.2\% of Spatial but only 73.0\% of Long. Short-horizon competence
survives in models far too small to chain the subgoals of long-horizon tasks,
suggesting the long suite is where LIBERO actually prices capacity.

\subsection{What survives seed averaging}
\label{sec:seeds}

\begin{table}[tb]
\caption{Ablations at the 1M scale, each cell a full 2{,}000-rollout
evaluation. Top: configurations re-trained with three seeds. Bottom: single-run deltas
(seed 1000) --- suggestive only against the $\pm$1-point seed band.}
\label{tab:ablations}
\centering
\footnotesize
\setlength{\tabcolsep}{3pt}
\begin{tabular}{@{}lrr@{}}
\toprule
change & avg $\pm$ sd ($n{=}3$) & $\Delta$ \\
\midrule
baseline: chunk 16, flow matching & $95.63 \pm 1.08$ & --- \\
action chunk $16 \to 8$ & $92.47 \pm 0.84$ & $\mathbf{-3.16}$ \\
action chunk $16 \to 32$ & $93.43 \pm 0.71$ & $\mathbf{-2.20}$ \\
vision $0.49\text{M} \to 0.13$M (same total) & $94.22 \pm 0.20$ & $\mathbf{-1.41}$ \\
flow matching $\to$ L1 regression & $95.97 \pm 0.80$ & $+0.34$ \\
\midrule
& ($n{=}1$) & \\
AdaLN-Zero $\to$ concatenation & & $-0.40$ \\
AdaLN-Zero $\to$ FiLM & & $-0.10$ \\
no distillation & & $+0.10$ \\
mixer $\to$ DiT self-attn.\ head ($+26\%$ params) & & $+0.05$ \\
\bottomrule
\end{tabular}
\end{table}

Re-training the configurations of Table~\ref{tab:ablations} (top) under three
training seeds first calibrates the instrument: \textbf{a single training
seed moves the 4-suite average by $\pm$1 point} (the baseline spans
94.60--96.75; its long-suite score alone spans 87.6--92.8). This is large enough to obscure the roughly one-point deltas common in single-run ablations.

Two choices survive the band. The first is \textbf{chunk length} ($-3.16$ at
chunk 8; $-2.20$ at chunk 32). Its
failure is
asymmetric and interpretable. Chunk 8 breaks the \emph{goal} suite (86--89\%
vs.\ 96.6--97.4\% at chunk 16) while leaving the long suite intact: goal
tasks pose different targets in one scene, so the policy must commit to a
branch, and short chunks let it dither between modes. Chunk 32 breaks the
\emph{long} suite (80--82\% vs.\ 87.6--92.8\%) while leaving goal intact:
multi-stage tasks require reacting at subgoal boundaries, and long open-loop
chunks cannot. \textbf{Chunk length chooses which suite you sacrifice}; 16
is the value that sacrifices neither.

The second is \textbf{vision allocation}: the starved 0.13M-vision configuration,
re-trained under three seeds after its single-seed delta ($-2.50$) proved
the last load-bearing $n{=}1$ number, replicates at $94.22 \pm 0.20$, a
$-1.41$-point mean difference, with the damage concentrated on the
long suite ($-4.6$). This is half the single-seed estimate, but the direction is consistent across seeds; notably the
starved model is far more seed-stable (sd 0.20 vs.\ 1.08) --- with less
vision capacity there is less to vary.

One widely assumed choice does \emph{not} survive: \textbf{flow matching is
indistinguishable from direct L1 regression}. The single-seed comparison
had regression $-1.70$ worse; over three seeds this reverses to $+0.34$.
Yet regression requires only one forward pass rather than ten Euler steps,
reducing inference time from 8.2 to 2.2\,ms per chunk on GPU and from 8.9
to 5.1\,ms on CPU at the 0.54M scale (Table~\ref{tab:speed}). Flow matching therefore provides no
measurable accuracy benefit under this benchmark and protocol, whereas
direct regression is \textbf{up to 3.8$\times$ faster}. The same conclusion
holds at 0.54M parameters, where retraining with the regression head yields
95.75\% versus 95.05\% for flow matching, well within the seed band despite
the noisier velocity field. We therefore ship the regression variant.

The single-run rows (Table~\ref{tab:ablations}, bottom) read accordingly:
conditioning style, distillation at 1M, and restoring self-attention
($+0.05$ at $+26\%$ parameters) are all far inside the band --- no measured
difference; the mixer head's value is that it reaches the same score with
26\% fewer parameters.

\subsection{Where to spend a fixed budget: the eyes, not the head}

A six-point sweep redistributing a fixed $\approx$1M budget between vision
encoder and action head is one-sided: every allocation giving vision
50--80\% of the budget lands in a flat 96.3--96.8 band, while starving
vision to 0.13M costs the seed-replicated $-1.41$ average and $-4.6$ long
of Section~\ref{sec:seeds} --- a loss no action-head capacity recovers.
During development the same effect appeared as a 10-point long-suite jump
(80.8$\to$91.0) when moving parameters from an attention head into the CNN
at constant total. The design rule for tiny policies is blunt:
\textbf{shrink the action head, not the eyes}.

\subsection{Inference-time strategy}
\label{sec:inference}

A comparison of four execution strategies on the same 0.54M checkpoint (25
episodes/task screens) shows two regularities. First, \textbf{execution
horizon dominates}: for every method, executing 1--2 actions per replan
clearly beats 4, which beats 8. Second, at horizon 1, averaging overlapping chunk predictions (temporal ensembling, 96.3\%) beats BID-style candidate selection (95.0\%), which selects the most coherent of 16 sampled chunks, and both outperform plain chunking (94.1\%); RTC-style soft inpainting \emph{hurts}
(91.8\%), over-constraining an already-noisy velocity field. The winner was
re-confirmed at 50 episodes/task and is the protocol behind
Table~\ref{tab:headline}.

Mode-seeking sampling behaves like a capacity-dependent corrector: initial-noise
temperature 0.85 adds $+0.2$ to $+2.25$ points on every sub-1M model ---
most where the model is weakest --- and nothing at 1M and above. Small flow
policies do not need a different architecture at inference; they need their
noisy velocity field averaged and their sampling sharpened.

\subsection{Efficiency: closing the loop on a CPU}

\begin{table}[tb]
\caption{Inference cost per action chunk (RTX 5080 Laptop GPU / 8 CPU
threads, batch 1).}
\label{tab:speed}
\centering
\small
\begin{tabular}{@{}lrrrr@{}}
\toprule
policy & params & GPU ms & CPU ms & VRAM \\
\midrule
MINERVA-0.5M & 0.54M & 8.2 & 8.9 & 0.03\,GB \\
\quad + L1 head & 0.54M & \textbf{2.2} & \textbf{5.1} & 0.03\,GB \\
MINERVA-1M   & 0.99M & 8.5 & 10.0 & 0.03\,GB \\
MINERVA-10M  & 9.66M & 11.1 & 17.8 & 0.08\,GB \\
SmolVLA      & 450M  & 118  & 1{,}010 & 0.97\,GB \\
\pizerofive{} & 4.1B & 196 & 12{,}781 & 9.36\,GB \\
\bottomrule
\end{tabular}
\end{table}

Table~\ref{tab:speed} reports wall-clock cost: on identical hardware the
0.54M model is 113$\times$ faster than SmolVLA and 1{,}400$\times$ faster
than \pizerofive{}.
At 8.9\,ms per chunk on eight laptop CPU threads --- $\approx$5\,ms with the
shipped regression head, which skips the ten-step ODE --- MINERVA-0.5M can
replan at every step of a $\sim$100\,Hz control loop with no GPU at all.
The replan-every-step protocol that maximizes success
(Section~\ref{sec:inference}) is therefore affordable on an embedded CPU;
for the VLA baselines it is not ($\sim$1\,Hz SmolVLA, 0.08\,Hz
\pizerofive{}). For integration into a larger
robot system, the entire perception-to-action module fits in the latency and
memory budget usually reserved for a single sensor driver --- and task-ID
conditioning is a natural interface for it: a planner that already knows
which skill to execute selects it by index, no language model in the loop.

\subsection{The embedding causally selects the task}
\label{sec:probe}
The headline result --- task IDs replace language at no cost --- is correlational. A permutation probe makes the role of task conditioning causal: we evaluate the released 1M checkpoint, with all weights fixed, while rewriting only its task-ID mapping. Rotating every task's ID within its suite collapses the average success rate from 96.75\% to \textbf{6.5\%}. Forcing every task to its suite's task-0 ID yields \textbf{16.2\%}. Spatial, Goal, and Long fall almost exactly to the 1-in-10 prediction (10.0\%, 10.0\%, and 8.8\%, respectively), under which only the task matching the forced ID succeeds. Object retains a larger 22--36\% residue, indicating that visually distinctive scenes can partially compensate for an incorrect task ID. These results show that task-ID conditioning provides the dominant task-selection signal, while visual context can disambiguate the task when the conditioning signal is insufficient. On standard LIBERO, instruction-following therefore reduces primarily to task \emph{selection}, with task identity conveyed mainly through the conditioning channel.

\subsection{Beyond 40 tasks: LIBERO-90}

To test the recipe on a larger closed task set, we train the 1M
architecture on LIBERO-90 (3{,}959 demonstrations, 90 tasks; one task has no
demonstrations). Task IDs are indices into the list of distinct instruction
strings, and the 90 tasks contain only 74 distinct instruction strings, so
some tasks necessarily share an ID. The result: \textbf{94.6\% average over 89
tasks} at 0.995M parameters --- standard-LIBERO-level success at
$2.25\times$ the task count, with no language encoder, no pretrained vision,
and here no distillation. Tasks sharing an instruction string with another
task score 93.6\% vs.\ 95.1\% for unique-ID tasks: vision supplies scene
context when the ID underdetermines it, completing the picture of
Section~\ref{sec:probe}. The failure tail is thin (worst tasks 50--60\%; no collapsed scene
family). Caveats: single seed, 10 episodes/task, evaluation pinned to the
renderer version matching this dataset's scenes.

\subsection{Robustness under LIBERO-Plus perturbations}

\begin{table}[tb]
\caption{LIBERO-Plus~\cite{liberoplus} success (\%) by perturbation factor
(stratified sample, 2 episodes/variant, identical variants per model).
Language variants are visually identical to base scenes and double as the
unperturbed baseline.}
\label{tab:plus}
\centering
\footnotesize
\setlength{\tabcolsep}{4pt}
\begin{tabular}{@{}lrrr@{}}
\toprule
factor & 0.54M & 0.99M & 4.89M \\
\midrule
language (= baseline) & 96.7 & 96.7 & 99.2 \\
new objects & 69.7 & 62.1 & 95.5 \\
sensor noise & 68.5 & 60.0 & 73.8 \\
robot init & 57.1 & 60.3 & 63.5 \\
object layout & 50.0 & 51.8 & 58.9 \\
camera & 11.5 & 18.5 & 34.6 \\
background & 9.3 & 4.7 & 7.0 \\
light & 1.1 & 2.2 & 6.5 \\
\midrule
\textbf{average} & \textbf{46.7} & \textbf{46.0} & \textbf{55.7} \\
(standard LIBERO) & (95.05) & (96.75) & (97.15) \\
\bottomrule
\end{tabular}
\end{table}

Finally we price what the 95\% headline does \emph{not} include, on
LIBERO-Plus~\cite{liberoplus} (10{,}030 perturbed variants of the 40 tasks
across 7 factors; each rephrased instruction is mapped to the corresponding base-task ID). In
Table~\ref{tab:plus} the 0.54M, 0.99M and 4.89M models fall to 46.7\%,
46.0\% and 55.7\%. Three findings: (1) language variants lose $\approx$0
points, so the drop is perturbation-driven, not a renderer artifact; (2)
robustness is not capacity-limited below 1M, and 5M buys $+10$ points,
almost all on semantic/geometric factors (new objects 62$\to$96); (3)
photometric robustness remains near zero across all tested scales (light $\leq$ 6.5\%): the scratch CNN is highly sensitive to appearance shifts, and no parameter
count in this range fixes it --- quantifying the memorization thesis.

\section{DISCUSSION AND LIMITATIONS}
\label{sec:limitations}

\textbf{What this says about the benchmark.}
MINERVA is deliberately incapable of language understanding, open-vocabulary
perception, or transfer to unseen tasks --- yet it loses only 0.75--2.4
points to state-of-the-art VLAs on standard LIBERO. Together with the
LIBERO-PRO perturbation analysis~\cite{liberopro}, the parameter-scaling
curve gives a quantitative reading: the standard evaluation is satisfiable by
$\sim$0.5M parameters of task-indexed visuomotor memorization --- and the
permutation probe of Section~\ref{sec:probe} shows that the task ID selects
the executed task, while the LIBERO-Plus audit prices exactly what the
memorization omits. Reported
LIBERO gains above $\sim$97\% are therefore unlikely to measure the
capabilities that motivate large models; perturbation-based
protocols~\cite{liberopro} are needed to see those. The training-seed band of Section~\ref{sec:seeds} sharpens this:
single-run gains of a point or less on this benchmark are comparable to
the observed seed-to-seed variation. Conversely, the long suite's sharp collapse below
0.5M is the one axis where standard LIBERO does price capacity.

\textbf{A capacity floor for deployment.}
The point of scaling down is not the small model itself but the measurement.
Generalist VLAs concentrate their value in open-world generalization; a
deployed system with a fixed task set does not pay for generalization it
will not use --- the argument deployment-efficient RL makes for training
constraints~\cite{bremen}, applied here to capacity. Our result gives one
concrete task set its empirical floor ($\approx$0.5M parameters, $\approx$1M with
long-horizon headroom), and the recipe --- train or reuse a larger model,
distill to the floor, execute with one forward pass --- is the pipeline an
integrator would run against their own task set. The floor is
task-set-specific: LIBERO-Plus suggests that robustness may require greater
capacity --- the 4.89M model substantially improves semantic robustness
over the sub-1M models --- and it will shift with environment, sensing and
embodiment. Measuring how this floor varies
--- a manipulation analogue of information-theoretic task-complexity
measures in RL~\cite{pic} --- is the research program this first step
opens.

\textbf{Limitations.}
The task-ID embedding cannot address a task it was not trained on; MINERVA is
a measurement instrument and a deployable component for closed task sets, not
a generalist. The five key ablation configurations are each evaluated across three training seeds,
but the headline models, the LIBERO-90 run, the remaining sweep points and
the distillation comparison are single runs inside a $\pm$1-point seed band,
and everything rests on one simulator and one evaluation seed. Robustness is
now measured rather than assumed: under LIBERO-Plus perturbation the models
keep roughly half their standard score, failing hardest on photometric
shifts --- a potentially addressable gap, e.g., via photometric augmentation,
that the 95\% headline conceals. Real-robot validation remains future work, though
the CPU latency and the SO-101-class hardware support in the underlying
framework~\cite{lerobot} make it direct.

\section{CONCLUSION}

We scaled a language-free, from-scratch manipulation policy down until
standard LIBERO broke it, and it did not break until half a million
parameters. A $\pm$1-point training-seed band frames every comparison;
only chunk length and vision allocation survive it, and one-pass regression
matches flow matching at lower inference cost and is therefore the
configuration we ship. A permutation probe
shows the embedding causally selects the task, the recipe holds on
LIBERO-90, and a LIBERO-Plus audit locates the robustness the headline does
not include. The outcome is a manipulation module that closes its control
loop in $\approx$5\,ms on a laptop CPU, and a calibration of what the
community's default benchmark actually measures --- a first step toward
task-specific capacity measurement: know a task set's empirical floor, then build or
distill to it.

\section*{ACKNOWLEDGMENT}

The authors thank the maintainers of LeRobot and LIBERO for the open
infrastructure this study builds on. This research also used computational resources of Miyabi provided through the Multidisciplinary Cooperative Research Program at the Center for Computational Sciences, University of Tsukuba.

\bibliographystyle{IEEEtran}
\bibliography{references}

\end{document}